\documentclass[article]{llncs}
\usepackage{epsfig}
\usepackage{graphicx}
\usepackage{amsfonts}
\usepackage{amsmath}
\usepackage{amssymb}
\usepackage{bbm}
\usepackage{multirow}
\usepackage{makecell}
\usepackage[pagebackref=false,breaklinks=true,colorlinks,bookmarks=false]{hyperref}

\usepackage{tikz}
\usepackage{comment}
\usepackage{amsmath,amssymb} 
\usepackage{color}

\usepackage[accsupp]{axessibility}  

\begin{document}
\pagestyle{headings}
\mainmatter
\def\ECCVSubNumber{7}  

\title{Efficient Visual Tracking via Hierarchical Cross-Attention Transformer} 

\titlerunning{Abbreviated paper title}
%
\author{Xin Chen\inst{1} \and
Ben Kang\inst{1} \and
Dong Wang\inst{1}\thanks{Corresponding author: Dr. Dong Wang, wdice@dlut.edu.cn} \and
Dongdong Li\inst{2} \and
Huchuan Lu\inst{1,3}}

\authorrunning{F. Author et al.}
%
\institute{School of Information and Communication Engineering, Dalian University of Technology, China \and
National Key Laboratory of Science and Technology on Automatic Target Recognition, National University of Defense Technology, China \and
Peng Cheng Laboratory
}
\maketitle

\begin{abstract}
In recent years, target tracking has made great progress in accuracy. This development is mainly attributed to powerful networks (such as transformers) and additional modules (such as online update and refinement modules). However, less attention has been paid to tracking speed. Most state-of-the-art trackers are satisfied with the real-time speed on powerful GPUs. However, practical applications necessitate higher requirements for tracking speed, especially when edge platforms with limited resources are used. In this work, we present an efficient tracking method via a hierarchical cross-attention transformer named HCAT. Our model runs about 195 $fps$ on GPU, 45 $fps$ on CPU, and 55 $fps$ on the edge AI platform of NVidia Jetson AGX Xavier. Experiments show that our HCAT achieves promising results on LaSOT, GOT-10k, TrackingNet, NFS, OTB100, UAV123, and VOT2020. Code and models are available at \href{https://github.com/chenxin-dlut/HCAT}{https://github.com/chenxin-dlut/HCAT}.

\end{abstract}

\section{Introduction}
\label{sec-int}

\begin{figure*}[!t]
\begin{minipage}[c]{0.39\textwidth}
   \caption{Speed and performance comparison on TrackingNet~\cite{trackingnet}. The horizontal axis is the model's speed on the edge AI platform of NVidia Jetson AGX Xavier. The vertical axis is the success (AUC) score. Following the VOT real-time setting~\cite{vot2020}, we set the real-time line to 20 $fps$. Our method achieves the best real-time result. We also apply our hierarchical cross-attention layer to TransT~\cite{TransT}, namely 
   TransT\_H. TransT\_H achieves the best result over all compared trackers. }
\end{minipage}
\begin{minipage}[c]{0.59\textwidth}
   \includegraphics[width=\textwidth]{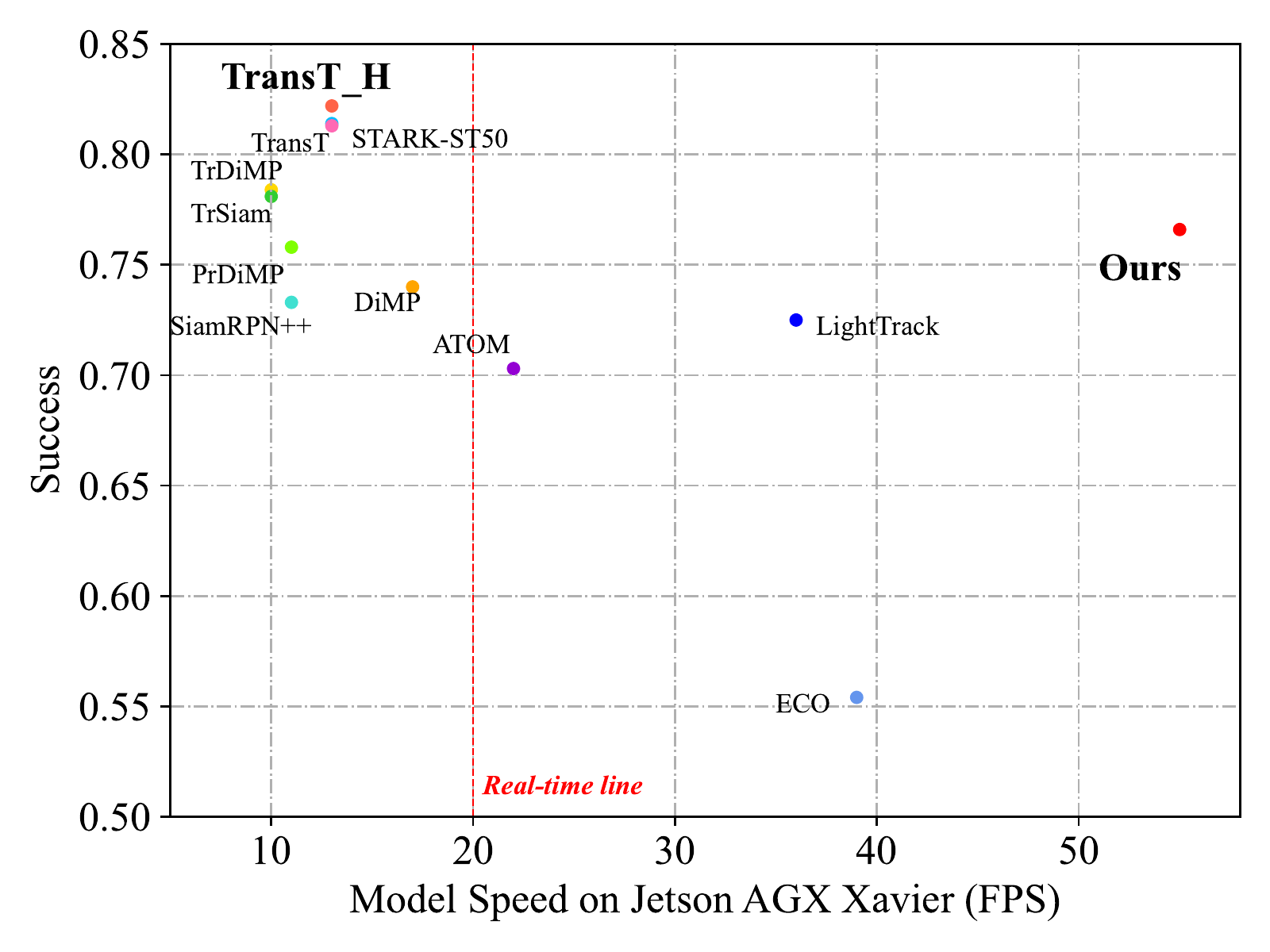}
\end{minipage}
\label{fig:trade}
\end{figure*}

Visual object tracking is a fundamental task in computer vision, in which the aim is to track an arbitrary object in a video given its initial location.
It is widely used in drone vision, autonomous driving, surveillance, and other fields.
Over the past few years, object tracking has made great progress owing to the development of deep networks~\cite{AlexNet,ResNet,googlenet,2017Attention}.
Most trackers aim to achieve high performance on datasets, but they ignore tracking speed and appear satisfied with real-time speed on powerful GPUs.
However, real-world application scenarios require trackers to function in real-time or even faster on edge devices, such as CPUs and NVidia Jetson devices.
However, most of the recent state-of-the-art trackers cannot achieve real-time speed on edge devices as shown in Fig.~\ref{fig:trade}, thus limiting their real-world applications.
In this work, we attempt to develop an accurate and extremely fast tracking algorithm.

Recently, transformer~\cite{2017Attention} has been successfully applied in many vision tasks \cite{ViT,liu2021swin,DETR}.
In the tracking field, transformer also boosts the performance~\cite{TransT,Wang_2021_Transformer,Stark}.
However, transformer entails high seriality and its computational amount is proportional to the square of the number of input tokens.
In this work, we design a more efficient transformer structure based on the variant transformer in TransT~\cite{TransT}.
There are two straightforward ways to speed up the transformer.
One approach is to reduce the calculation amount of the operation layers, such as the linear transformation layer, convolution layer, and attention layer.
On platforms with limited computing power, the model's speed can be improved in this way.
However, reducing the computational amount of operation layers does not always effectively improve the model's speed because devices have the ability of parallel computing.
For example, we test the latency of feed-forward network FFN~\cite{2017Attention,DETR} commonly used in transformer under different calculation amounts, as shown in Table~\ref{tab-ffn}.
FFN1 has $4\times$ less calculation amount than FFN2, but its latency is not less on GPU and only $2.7\times$ less on CPU.
Another way is to reduce the seriality of the transformer.
This approach involves directly reducing the number of layers of the network, which can bring considerably speedup on both edge platforms and powerful GPUs.
However, reducing the seriality causes the network to be shallow and it weakens the feature expression ability.
So the target is to keep the efficient layers and discard the relatively inefficient layers.

\begin{table*}[t]
\setlength{\abovecaptionskip}{0cm} 
\setlength{\belowcaptionskip}{-0.2cm}
\caption{Running speed of FFN with different calculation amounts.}
\label{tab-ffn}
\begin{center}
\resizebox{0.95\textwidth}{!}{
\setlength{\tabcolsep}{4mm}{
\begin{tabular}{c ccc c}
\hline
&Input Size
&MFlops
&GPU latency (ms)
&CPU latency (ms)\\
\hline
FFN1 &$8 \times 8 \times 256$ &67.1 &0.15 &0.45\\
FFN2 &$16 \times 16 \times 256$ &268.4 &0.15 &1.22\\
\hline
\end{tabular}}}
\end{center}
\end{table*}

In this work, to speed up the transformer, we first design the \textbf{H}ierarchical \textbf{C}ross-\textbf{A}ttention (HCA) transformer structure to reduce the seriality and improve the model's representation ability. We employ a full cross-attention structure to improve efficiency and a hierarchical connection method to deepen the network, subsequently enhancing the representation ability. Then, we design the \textbf{F}eature \textbf{S}parsification (FS) module to sparse the template feautures and reduce the computational amount of the transformer. On the basis of these two modules, we propose an efficient visual tracking method named HCAT.

Our contributions can be summarized as follows.
\begin{itemize}
    \setlength{\itemsep}{0pt}
    \setlength{\parsep}{0pt}
    \setlength{\parskip}{0pt}
    \item We propose a novel hierarchical cross-attention transformer that employs a full cross-attention structure to improve efficiency and a hierarchical connection method to enhance
    the representation ability under a limited number of operation layers.
	\item We propose a feature sparsification module to sparse the template features and reduce the computational amount of the transformer without affecting performance.
	\item The feature sparsification module and the hierarchical cross-attention transformer form a new feature fusion network. We combine the feature fusion network with the backbone network and prediction head to develop a new efficient tracker named HCAT.
    \item Our HCAT has an extremely fast speed. The PyTorch model runs at 195 $fps$ on GPU, 45 $fps$ on CPU, and 55 $fps$ on the edge AI platform of NVidia Jetson AGX Xavier. The ONNX model runs at 589 $fps$ on GPU, 90 $fps$ on CPU, and 127 $fps$ on NVidia Jetson AGX Xavier. Numerous experimental results on many benchmarks show that the proposed tracker performs considerably better than the state-of-the-art high-speed algorithms.
\end{itemize}

\section{Related Work}
\label{sec-relate}

{\noindent \textbf{Visual Object Tracking. }}
In recent years, Siamese trackers have become popular in visual object tracking.
SiamFC~\cite{SiameseFC} and SINT~\cite{SINT}, the pioneering works, combine naive feature correspondence by using the Siamese framework.
Since then, SiamRPN~\cite{SiameseRPN} has used the RPN~\cite{FasterRCNN} into the Siamese tracking framework for precise bounding box estimation.
Many improvements have been achieved to boost tracking performance, such as developing additional branches \cite{SiamMask,Alpha-Refine}, employing deeper architectures~\cite{SiamRPNplusplus,Deeper-wider-SiamRPN,Deform_siam}, exploiting the anchor-free mechanism~\cite{SiamFC++,SiamBAN,SiamCAR,Ocean}, and so on.
Online trackers~\cite{MDNet,ECO,ATOM,DiMP,Ocean} improve robustness by employing online updating modules. 
However, due to complex designs, most previous Siamese trackers or online trackers only achieve real-time speed on powerful GPUs, but they hardly achieve real-time speed on edge platforms.
The Siamese tracking framework can be divided into three parts: the backbone network, the feature fusion network, and the prediction head.
In this work, aiming to develop an extremely fast tracker, we employ a simple backbone network and a simple prediction head network while carefully designing an efficient feature fusion network. Our method can achieve real-time speed on edge platforms.

{\noindent \textbf{Transformer in Tracking. }}
Vaswani \emph{et al.} introduced transformer~\cite{2017Attention} in machine translation.
Transformer is composed of attention-based encoders and decoders.
Attention mechanism has achieved remarkable results in many tasks because of its ability to integrate global information and less inductive bias.
DETR and ViT~\cite{DETR,ViT} were the first to apply transformer into computer vision field.
After that, transformer has been successfully applied in a number of computer vision tasks.
In object tracking, transformer has brought great performance improvement.
TransT~\cite{TransT} exploits the core idea of transformer and develops a new feature fusion network to fuse template and search region features by using cross-attention rather than correlation operation.
TMT~\cite{Wang_2021_Transformer} uses transformer as a feature enhancement module and combines it with SiamRPN~\cite{SiameseRPN} and DiMP~\cite{DiMP}.
STARK~\cite{Stark} employs the transformer by inputting the concatenated template and search region features. 
DualTFR and SwinTrack~\cite{DualTFR,liu2021swin} use transformer as the backbone network.
In this work, we develop an efficient feature fusion network based on transformer.
We choose TransT as our baseline because it is relatively simple, does not require additional modules, and has a good speed on GPUs.
We develop a feature sparsification module and a hierarchical cross-attention transformer to enable the tracker to achieve real-time speed on edge platforms with good performance.

{\noindent \textbf{Efficient Tracking Network. }}
More and more applications that use tracking algorithms have been implemented, including unmanned driving, UAV vision and robot vision.
In real-world applications, trackers usually run on the edge platforms with limited computing power. 
However, most of the current state-of-the-art trackers only run fast on powerful GPUs.
The demand for efficient tracking network is urgent.
Unfortunately, in recent years, researchers have hardly paid attention to tracker's speed on edge devices.
ATOM~\cite{ATOM} and ECO~\cite{ECO} can achieve real-time speed on NVidia Jetson AGX Xavier (AGX); however, compared with the popular tracking algorithms (such as PrDiMP and SiamRPN++), their performance is poor.
LightTrack~\cite{yan2021lighttrack} is the latest lightweight tracking algorithm that uses NAS to search networks, and it entails low computational amount and relatively high performance.
However, a gap exists between the calculation amount and the model's real speed, as described in Section~\ref{sec-int}. 
LightTrack can achieve real-time speed on CPU and AGX; however, on powerful GPUs, the speed is not extremely fast.
We hope our tracker can achieve real-time speed on AGX and CPUs, and acheive extremely fast speed on powerful GPUs.
In this work, our tracker's PyTorch model runs at 195 $fps$ on GPU, 45 $fps$ on CPU, and 55 $fps$ on AGX. Our ONNX model runs at 589 $fps$ on GPU, 90 $fps$ on CPU, and 127 $fps$ on AGX.

\section{Method}
\label{sec-method}

\begin{figure*}[!t]
\begin{center}
\includegraphics[width=1\linewidth]{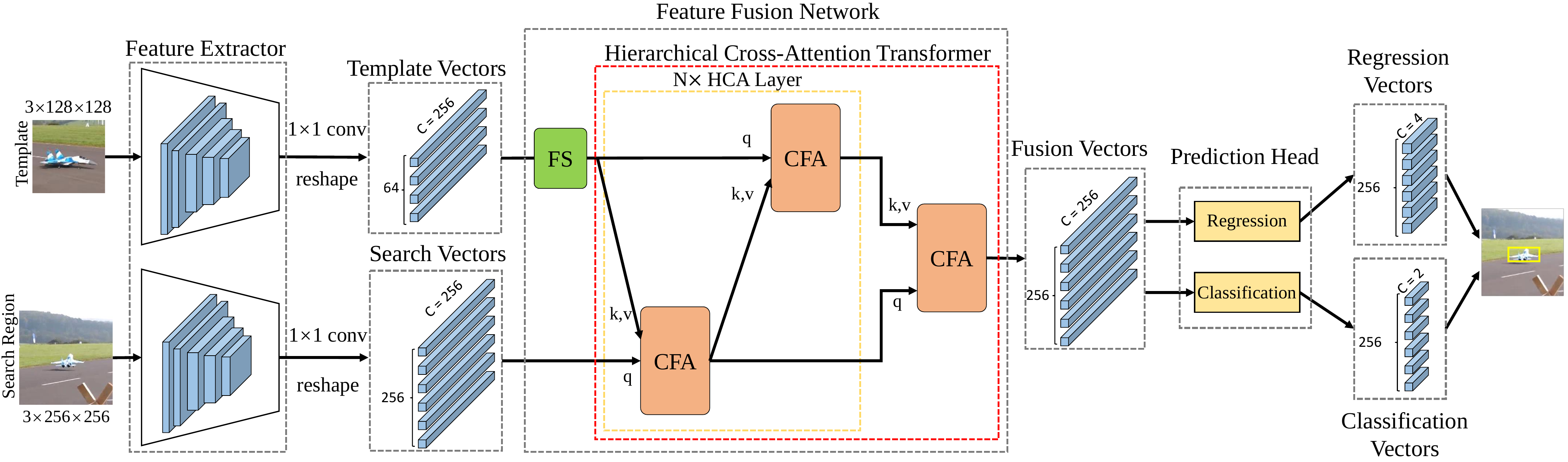}
\end{center}
   \caption{Architecture of our HCAT framework. This framework contains three components: 
   feature extraction backbone, feature fusion network, and prediction head. The feature fusion network consists of the feature sparsification module (denoted as FS) and the hierarchical cross-attention transformer.}
\label{fig:framework}
\end{figure*}

This section describes our method for HCAT. 
As shown in Fig.~\ref{fig:framework}, our method consists of the feature extraction network, the feature fusion network, and the prediction head. 
The feature extraction network extracts the features of the search region patch and the template patch. 
The feature fusion network fuses the search region features and the template features by using our hierarchical cross-attention transformer. 
Before feature fusion, we use the feature sparsification (FS) module to sparse the template feature vectors in the spatial dimension.
After feature fusion, we use the prediction head to perform bounding box regression and binary classification on the fusion feature vectors. 
Finally, we select the best bounding box according to the confidence score.
In this section, we introduce the details of each part of HCAT, introduce our hierarchical cross-attention transformer and feature sparsification module, and provide analysis. 

\subsection{Overall Architecture}

{\noindent \textbf{Feature Extraction. }}
Similar to most Siamese trackers~\cite{SiameseFC,SiameseRPN,SiamRPNplusplus}, we take the template image patch ($\mathbf{Z} \in {\mathbb{R}}^{3 \times {H_{z0}} \times {W_{z0}}}$) and the search region image patch ($\mathbf{X} \in {\mathbb{R}}^{3 \times {H_{x0}} \times {W_{x0}}}$) as the inputs of our HCAT.
The template patch is obtained by expanding the initial target bounding box outward by twice the side length, and the search region patch is obtained by expanding the bounding box of the previous frame's target by four times the side length.
The search region patch and template patch are reshaped to a square, and then the backbone network extracts their features. 
We use a modified version of ResNet18~\cite{ResNet} as the backbone network. 
Specifically, we remove the last stage of ResNet18. 
In contrast to previous Siamese trackers~\cite{SiamRPNplusplus,TransT}, we do not modify the stride; hence, the backbone's stride remains 16 rather than 8.
The large backbone stride reduces the feature's resolution, thus further reducing the computational cost. 
In this manner, the speed gain is not obvious on the powerful GPU, but it is obvious on edge platforms, as shown in Table~\ref{tab-stride}.
Finally, we use a $1 \times 1$ convolution layer to transform the channel dimension of the backbone features and flatten the features in the spatial dimension. 
Then, we obtain two sets of feature vectors: the template vectors ${\mathbf{F}_z} \in {\mathbb{R}}^{C \times {H_z}{W_z}}$ and the search vectors ${\mathbf{F}_x} \in {\mathbb{R}}^{C \times {H_x}{W_x}}$.
$H_z,W_z = {\frac{H_{z0}}{16}},{\frac{W_{z0}}{16}}$, $H_x,W_x = {\frac{H_{x0}}{16}},{\frac{W_{x0}}{16}}$, and $C = 256$.

\begin{table*}[t]
\setlength{\abovecaptionskip}{0cm} 
\setlength{\belowcaptionskip}{-0.2cm}
\caption{Model speed with different backbone strides.}
\label{tab-stride}
\begin{center}
\resizebox{0.95\textwidth}{!}{
\setlength{\tabcolsep}{6mm}{
\begin{tabular}{ccc c}
\hline
stride
&GPU speed ($fps$)
&CPU speed ($fps$)
&AGX speed ($fps$)\\
\hline
8 &178 &20 &22\\
16 &195 &45 &55\\
\hline
\end{tabular}}}
\end{center}
\end{table*}

{\noindent \textbf{Feature Fusion Network. }}
We design a new feature fusion network to fuse the features of the template and the search region.
As shown in Fig.~\ref{fig:framework}, the feature fusion network is composed of the FS module and the hierarchical cross-attention transformer.
First, the FS module sparses the template feature vector in the spatial dimension.
It reduces the number of template vectors to $S$, obtaining the sparse template vectors ${\mathbf{F}_{zs}} \in {\mathbb{R}}^{C \times {S}}$. 
Then, the hierarchical cross-attention transformer fuses the sparse template vectors and the search vectors.
The hierarchical cross-attention transformer is inspired by the feature fusion network proposed by TransT~\cite{TransT}. 
We employ the basic unit CFA in TransT, which is a residual structure based on the cross-attention layer and linear layer. 
CFA can fuse two sets of input features.
Two CFAs form a HCA layer, as shown in the yellow dotted box in  Fig.~\ref{fig:framework}.
In contrast to the method for TransT, we adopt the full cross-attention structure instead of employing self-attention layers.
Because the performance gain brought by self-attention layers is relatively small, but it requires extensive running time.
In TransT, two CFAs in the same feature fusion layer are juxtaposed but uncorrelated with each other.
Instead of juxtaposing two CFAs, we combine them in a hierarchical manner, i.e., the template branch's CFA receives the output of the search branch's CFA of the same layer as the key and value.
Hence, the cross-attention utilizes the more accurate features without incurring any additional computational cost. Under the limited number of operation layers, the network is deepened.
The HCA layer repeats $N$ times (here, $N=2$ by default). 
Then, an additional CFA decodes the final fusion feature vectors ${\mathbf{F}} \in {\mathbb{R}}^{C \times {H_x}{W_x}}$.
The details of the FS module and the hierarchical cross-attention transformer are introduced in Section~\ref{sec-FS} and Section~\ref{sec-hcat}.

{\noindent \textbf{Prediction Head Network. }}
The prediction head consists of the regression head and classification head, which are both a three-layered perceptron with a hidden dimension ($d=256$) and the $\rm{ReLU}$ activation function. 
The regression head directly outputs the normalized coordinates on each fusion vector, resulting in a total of ${H_x}{W_x}$ bounding boxes. 
The classification head performs binary (foreground/background) classification on each fusion vector. 
The classification head generates ${H_x}{W_x}$ classification scores corresponding to the bounding boxes.

\subsection{Feature Sparsification Module}
\label{sec-FS}
As shown in Fig.~\ref{fig:framework}, before the hierarchical cross-attention transformer, we use the FS module to sparse the template features, thus reducing the calculation amount of subsequent layers, especially the attention layers.
The FS module reduces the number of template vectors from ${H_z}{W_z}$ to $S$, and the subsequent linear layer's calculation are reduced accordingly.
For attention layer, apart from internal linear transformation, the attention mechanism is calculated as
\begin{equation}
    {\rm{Attention}}(\mathbf{Q},\mathbf{K},\mathbf{V})
    = {\rm{softmax}}(\frac{\mathbf{Q}\mathbf{K}^\top}{\sqrt{d}})\mathbf{V}, 
\label{eq-att}
\end{equation}
where ${\mathbf{Q}} \in {\mathbb{R}}^{d \times n_q}$ and ${\mathbf{K}},{\mathbf{V}} \in {\mathbb{R}}^{d \times n_k}$.  
The calculation amount of attention is given by $2d{n_q}{n_k}$; the calculation amount of multi-head attention is processed in the same manner.
However, efficiency of attention is criticized because its computational amount is proportional to the product of the two spatial dimensions $n_q$ and $n_k$.
Without the FS module in our method, the attention mechanism takes ${\mathbf{F}_z} \in {\mathbb{R}}^{C \times {H_z}{W_z}}$ and ${\mathbf{F}_x} \in {\mathbb{R}}^{C \times {H_x}{W_x}}$ as the inputs, resulting in the calculation amount of $2{H_z}{W_z}{H_x}{W_x}C$. 
The direct approach of minimizing the computational amount is to reduce the number of template vectors or search vectors.
We find that reducing the number of search vectors has a large impact on performance; however, under certain suitable settings, reducing the number of template vectors has a negligible impact on performance.
This finding can be attributed to the network requiring a classification and regression of the target on the search region.
The fine-grained appearance information and location information in search vectors are important for accurate predictions.
However, template features are relatively redundant as the network only needs to refer to it to know which target to track.
Reducing the number of search vectors or template vectors in equal proportions has the same impact on the computational amount of attention; thus we choose to sparse the template vectors.
Our FS module reduces the number of template vectors from ${H_z}{W_z}$ to $S$, and the computational amount of attention is reduced to $2{S}{H_x}{W_x}C$, where $S$ is a constant, which is much smaller than ${H_z}{W_z}$.
With our default settings of $S=16$ and ${H_z}{W_z}=64$, the computational amount is reduced by a quarter.
On powerful GPU, computational amount is not the bottleneck restricting the model's speed.
Therefore, this reduction cannot speed up the model on powerful GPUs, but it can substantially speed up the model on edge platforms.
The detail structure of the FS module is shown in Fig.~\ref{fig:detail} (a). 
The template vectors are the feature vectors ${\mathbf{F}_z} \in {\mathbb{R}}^{C \times {H_z}{W_z}}$ generated by backbone. 
The sparse vectors ${\mathbf{F}_s} \in {\mathbb{R}}^{C \times {S}}$ are the general features learned during training. 
The FS module is a residual structure based on the multi-head cross-attention layer.
Multi-head cross-attention takes the sparse vectors as query and the template vectors as key and value.
It extracts template vectors with reference to sparse vectors and outputs the sparse feature vectors.  
Then the sparse feature vectors are added to the sparse vectors ${\mathbf{F}_s}$ to obtain the final sparse template vectors ${\mathbf{F}_{zs}} \in {\mathbb{R}}^{C \times {S}}$.
The mechanism of the FS module can be summarized as
\begin{equation}
\begin{split}
\label{equation:FS}
\mathbf{F}_{zs} = \mathbf{F}_s + {\rm MHCA}(\mathbf{F}_s,\mathbf{F}_z + \mathbf{P}_z,\mathbf{F}_z)
\end{split}, 
\end{equation}
where $\mathbf{P}_z \in \mathbb{R}^{C \times {H_z}{W_z}}$ denotes the spatial positional encoding, which is generated by a sine function. ${\rm MHCA}(.,.,.)$ is the multi-head cross-attention mechanism.
We find that such a simple design achieves good results.

\subsection{Hierarchical Cross-Attention Transformer}
\label{sec-hcat}

\begin{figure*}[!h]
\begin{center}
\resizebox{\linewidth}{!}{
\begin{tabular}{c@{}c}
\includegraphics[width=0.40\linewidth]{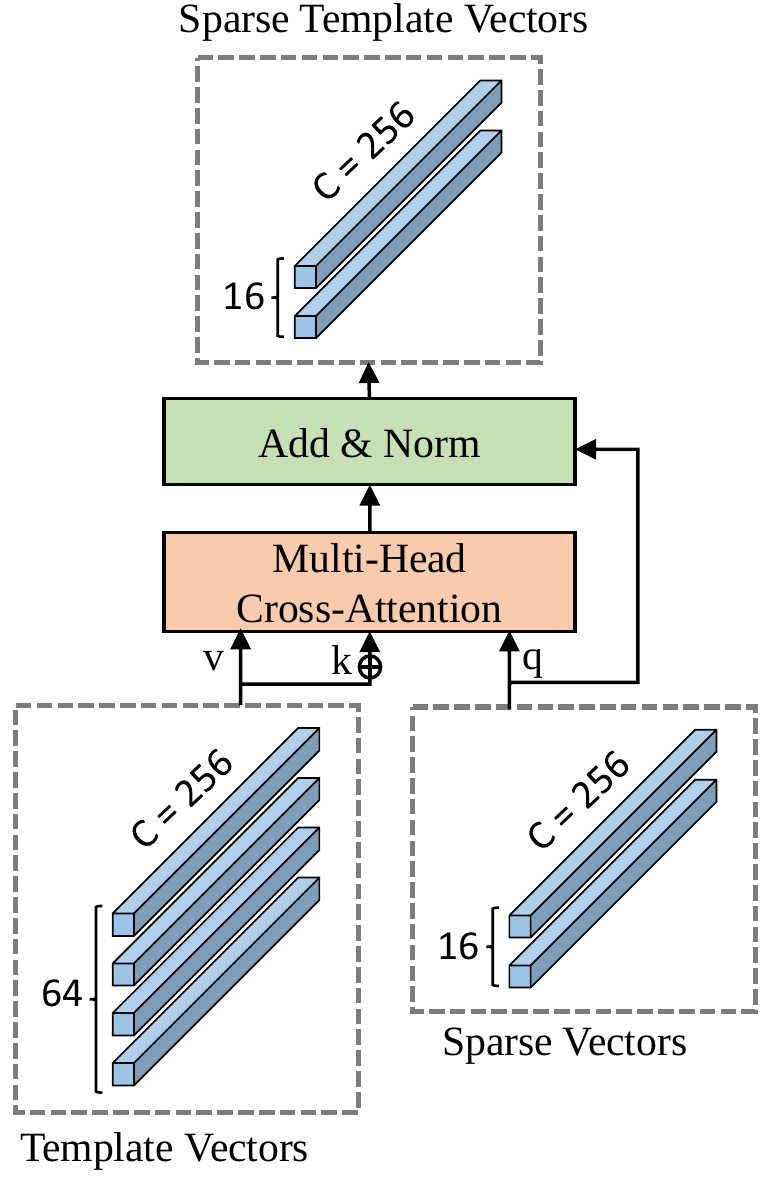} \ &
\includegraphics[width=0.85\linewidth]{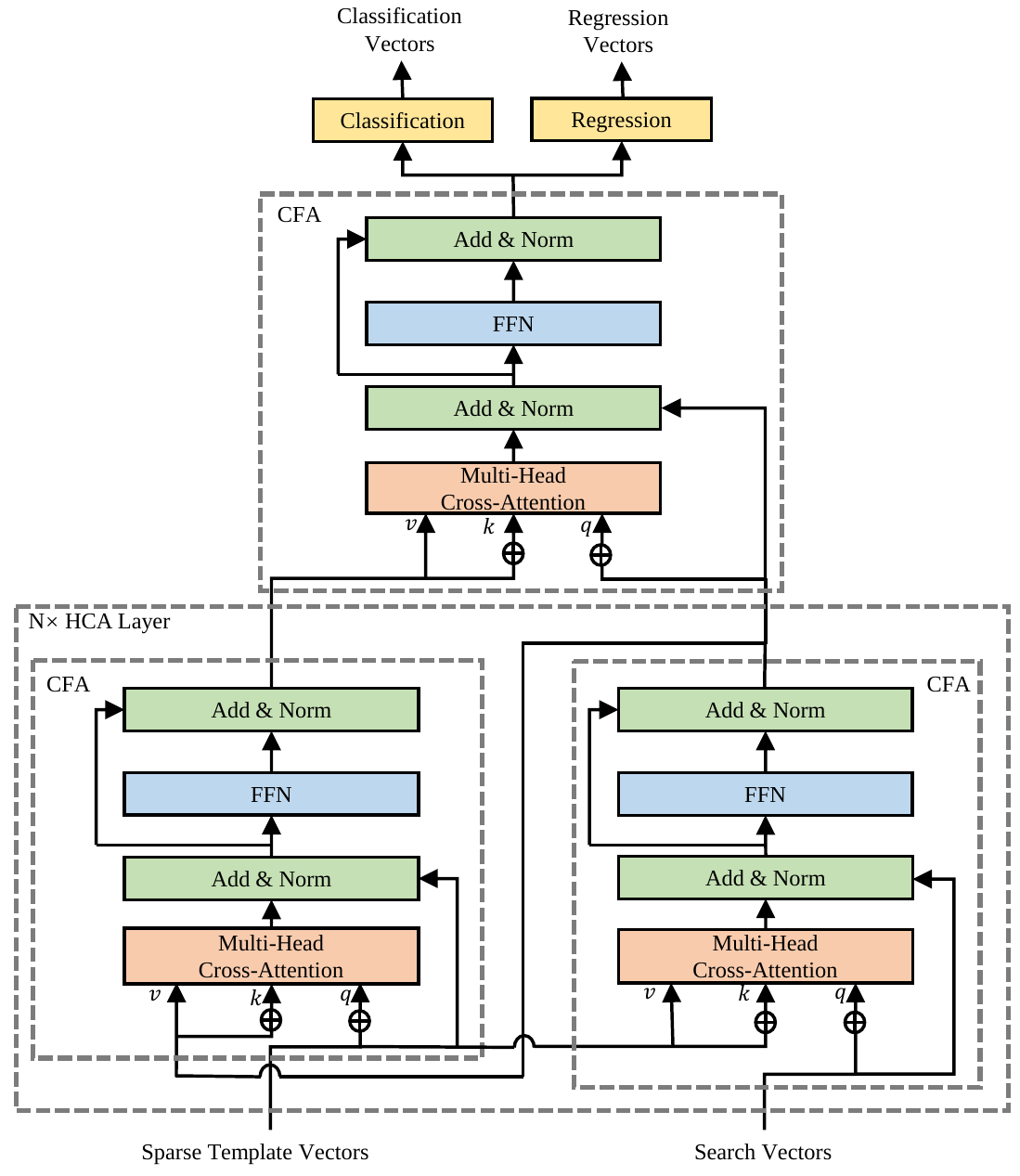} \\
(a) Feature Sparsification Module &(b) Hierarchical Cross-Attention Transformer\\
\end{tabular}}
\end{center}
\caption{Detailed architectures of our feature sparsification module and hierarchical cross-attention transformer. \textcircled{+} denotes adding spatial positional encoding. }
\label{fig:detail}
\end{figure*}

The detailed architecture of our hierarchical cross-attention transformer is shown in Fig.~\ref{fig:detail} (b).
This structure is based on the basic module of the CFA used in TransT.
CFA is a residual module based on multi-head cross-attention and FFN, as shown in the dotted box.
FFN is a feed-forward network that consists of two linear transformation layers with a $\rm{ReLU}$ activation function in between.
CFA is formulated as
\begin{equation}
\begin{array}{c}
\rm{CFA}(\bf{Q},\bf{K},\bf{V})={\bf{X}}_{CF},\\
{{\bf{X}}_{CF}} = {\widetilde{\bf{X}}_{CF}} + {\rm{FFN}}\left( {{{\widetilde{\bf{X}}}_{CF}}} \right), \\
{\widetilde{\bf{X}}_{CF}} = {{\bf{Q}}} + {\rm{MHCA}}\left( {{{\bf{Q}}} + {{\bf{P}}_q},{{\bf{K}}} + {{\bf{P}}_{kv}},{{\bf{V}}}} \right), 
\end{array}
\label{eq-cfa}
\end{equation}
where $\mathbf{Q} \in \mathbb{R}^{d \times n_q}$ is the $q$ input, $\mathbf{K}$ and $\mathbf{V} \in \mathbb{R}^{d \times n_{kv}}$ are the $k$ and $v$ inputs, where $\mathbf{K}=\mathbf{V}$ in our method. $\mathbf{P}_q \in \mathbb{R}^{d \times n_q}$ and $\mathbf{P}_{kv} \in \mathbb{R}^{d \times n_{kv}}$ are the spatial positional encodings.  
In our method, two CFAs form an HCA layer. 
The two CFAs in same layer are hierarchical rather than juxtaposed as in TransT.
Specifically, the template branch's CFA receives the output of the search branch's CFA of the same layer as the key and value rather than they do not affect each other like TransT.
The HCA layer can be formulated as
\begin{equation}
\begin{split}
\label{equation:HCA}
\mathbf{F}_{x}^{(l)} = {\rm CFA}(\mathbf{F}_{x}^{(l-1)} ,\mathbf{F}_{zs}^{(l-1)},\mathbf{F}_{zs}^{(l-1)}),\\
\mathbf{F}_{zs}^{(l)} = {\rm CFA}(\mathbf{F}_{zs}^{(l-1)} ,\mathbf{F}_{x}^{(l)},\mathbf{F}_{x}^{(l)}),
\end{split}, 
\end{equation}
where $(l-1)$ represents the previous layer, and $(l)$ represents the current layer.
In this manner, the CFA utilizes the more accurate features, which is equivalent to deepening the network without introducing additional layers.
Given the different sizes of the inputs, the two CFAs at the same layer cannot be operated in parallel in TransT and our method.
Therefore, the hierarchical connection method does not bring any additional running time cost.
We expect our model to achieve excellent speed not only on edge platforms but also powerful GPUs. 
As discussed in Section~\ref{sec-int} and Section~\ref{sec-FS}, owing to the powerful parallel computing capability of the GPU, the calculation amount of operation layers is not the bottleneck restricting the speed.
Minimizing the number of serial layers is more effective than reducing the computational amount of operation layers.
Therefore, we only use two HCA layers, avoid using any self-attention layer and instead adopt the full hierarchical cross-attention design.
In experiments, we find that the self-attention layer brings minimal performance benefit, and it requires a high running time.
These settings cause the network shallow, thus affecting the performance, but the hierarchical structure plays a mitigating role.

\subsection{Training Loss}
The prediction head receives $H_xW_x$ search vectors and generate $H_xW_x$ prediction results.
Each result includes a bounding box and a binary classification result. 
The prediction results whose corresponding positions in the ground-truth bounding box are assigned as positive samples, whereas the others are assigned as negative samples.
The class label of the positive samples is foreground and the class label of the negative samples is background.
All samples contribute to classification loss.
The regression label is the normalized bounding box coordinate.
Only the positive samples contribute to regression loss, which enable the feature vectors to predict the targets at their corresponding position.
Subsequently, the distractors can be filtered through the position information in online tracking.
For classification, we employ the binary cross-entropy loss. 
To balance the positive and negative samples, we reduce the classification loss contributed by the negative samples by a factor of 16.
For regression, we combine the $\ell_1$-norm loss and generalized IoU loss~\cite{GIoU} linearly.

\section{Experiments}
\label{sec-exp}

\subsection{Implementation Details}

{\noindent \textbf{Offline Training. }}
Our models are trained on the training set of LaSOT~\cite{LaSOT}, GOT-10k~\cite{GOT10K}, COCO~\cite{COCO}, and TrackingNet~\cite{trackingnet}. 
A search region patch and a template patch form a sampling pair.
For the video datasets of LaSOT, GOT-10k, and TrackingNet, we randomly take two frames in the same video sequence with an interval of no more than 100 frames to generate the template patch and search region patch.
For the image dataset of COCO, we transform the original image to generate sampling pairs.
We employ some normal data augmentation, such as position and scale transformation.
The search region patches are resized to $256 \times 256 \times 3$, and the template patches are resized to $128 \times 128 \times 3$.
The parameters of the backbone network are initialized with ResNet18~\cite{ResNet} pretrained on ImageNet~\cite{ImageNet}, other parameters are initialized with the Xavier init~\cite{Xavier}.
The models are trained with the AdamW~\cite{AdamW} optimizer. 
The learning rate of the backbone network is 1e-5, and that of the other parameters is 1e-4.
The weight decay is 1e-4.
We use 8 NVidia RTX 3090 GPUs to train our network for 500 epochs with a batch size 128.
Each epoch contains 60,000 sampling pairs.
The learning rate is decayed by $10 \times$ at epoch 400.

{\noindent \textbf{Online Tracking. }}
In online tracking, the network predicts $16 \times 16$ results with classification scores.
We employ window penalty~\cite{SiamRPNplusplus} to adjust the confidence score.
Specifically, we penalize the confidence scores far from the center of the search region by applying a Hanning window.
The window penalty is used to filter out the distractors.

\subsection{Evaluation on TrackingNet, LaSOT, and GOT-10k Datasets}

We compare our HCAT with state-of-the-art trackers on the three large-scale benchmarks of TrackingNet~\cite{trackingnet}, LaSOT~\cite{LaSOT}, and GOT-10k~\cite{GOT10K}.
We test the speed of the trackers' model on three platforms: GPU, CPU, and NVidia Jetson AGX Xavier (AGX).
Our GPU is the powerful NVidia Titan RTX, our CPU is Intel(R) Core(TM) i9-9900K @ 3.60Hz, and the AGX is an edge AI platform.
We only test the speeds of the models and exclude the pre-processing and post-processing of the image and results during tracking because the different implementations of these parts affect the tracking speed, and we do not want them to conceal the real speed of the models.
For the online update trackers (such as TrDiMP, PrDiMP, ATOM), we turn off their online update module, so their real speed is slower than our test speed.
We divide the trackers into two categories according to the running speed of the model.  
Specifically, following the VOT real-time setting~\cite{vot2020}, we set the real-time line to 20 $fps$.
The methods that can reach 20 $fps$ on our CPU or AGX are classified as real-time methods, whereas those that cannot reach 20 $fps$ on neither our CPU nor AGX are classified as non-real-time methods.
The detailed results are shown in Table~\ref{tab-sota}.

{\noindent \textbf{Speed. }}
The speeds on the three platforms are shown in Table~\ref{tab-sota}. AGX indicates NVidia Jetson AGX Xavier.
Our model's speed is 195 $fps$ on GPU, which is second only to ECO~\cite{ECO}, and it is 52\% higher than the recent lightweight tracker LightTrack~\cite{yan2021lighttrack}.
On the edge platforms, our model achieves 45 $fps$ on CPU (10\% higher than the second fastest tracker LightTrack) and 55 $fps$ on AGX (53\% higher than LightTrack).
In addition, after using ONNX to accelerate, our model's speed can reach 589 $fps$ on GPU, 90 $fps$ on CPU, and 127 $fps$ on AGX.
More results about the ONNX model are introduced in Section~\ref{sec-ablation}.

\begin{table*}[!t]
\setlength{\abovecaptionskip}{0cm} 
\setlength{\belowcaptionskip}{-0.2cm}
\caption{State-of-the-art comparison on TrackingNet, LaSOT, and GOT-10k benchmarks. The best real-time results 
are shown in \textbf{\textcolor{red}{red}} fonts, and the best non-real-time results are shown in \textbf{\textcolor{blue}{blue}} fonts.}
\label{tab-sota}
\begin{center}
\resizebox{\linewidth}{!}{
\begin{tabular}{cc|cc cc c|c ccc ccc}
\hline
& &\multicolumn{5}{c|}{Real-time} &\multicolumn{7}{c}{Non-real-time} \\
\hline
& &Ours 
&E.T.Track\cite{ETTrack}
&LightTrack\cite{yan2021lighttrack}
&ATOM\cite{ATOM} &ECO\cite{ECO}
&STARK-ST50\cite{Stark}
&TransT\cite{TransT}
&TrDimp\cite{Wang_2021_Transformer} &TrSiam\cite{Wang_2021_Transformer}
&PrDiMP\cite{PrDiMP} &DiMP\cite{DiMP} &SiamRPN++\cite{SiamRPNplusplus}  \\
\hline
\multirow{3}{*}{TrackingNet} &AUC &\textbf{\textcolor{red}{76.6}} &75.0 &72.5 &70.3 &55.4 &81.3 &\textbf{\textcolor{blue}{81.4}} &78.4 &78.1 &75.8 &74.0 &73.3 \\
&P$_{Norm}$ &\textbf{\textcolor{red}{82.6}} &80.3 &77.8  &77.1 &61.8 &86.1 &\textbf{\textcolor{blue}{86.7}} &83.3 &82.9 &81.6 &80.1 &80.0 \\
&P &\textbf{\textcolor{red}{72.9}} &70.6 &69.5 &64.8 &49.2 &- &\textbf{\textcolor{blue}{80.3}} &73.1 &72.7 &70.4 &68.7 &69.4 \\
\hline
\multirow{3}{*}{LaSOT} &AUC &\textbf{\textcolor{red}{59.3}} &59.1 &53.8 &51.5 &32.4 &\textbf{\textcolor{blue}{66.6}} &64.9 &63.9 &62.4 &59.8 &56.9 &49.6 \\
&P$_{Norm}$ &\textbf{\textcolor{red}{68.7}} &- &- &57.6 &33.8 &- &\textbf{\textcolor{blue}{73.8}} &- &- &68.8 &65.0 &56.9 \\
&P &\textbf{\textcolor{red}{61.0}} &- &53.7 &50.5 &30.1 &- &\textbf{\textcolor{blue}{69.0}} &61.4 &60.0 &60.8 &56.7 &49.1 \\
\hline
\multirow{3}{*}{GOT-10k} &AO &\textbf{\textcolor{red}{65.1}} &- &61.1 &55.6 &31.6 &68.0 &\textbf{\textcolor{blue}{72.3}} &68.8 &67.3 &63.4 &61.1 &51.7 \\
&SR$_{0.5}$ &\textbf{\textcolor{red}{76.5}} &- &71.0 &63.4 &30.9 &77.7 &\textbf{\textcolor{blue}{82.4}} &80.5 &78.7 &73.8 &71.7 &61.6 \\
&SR$_{0.75}$ &\textbf{\textcolor{red}{56.7}} &- &- &40.2 &11.1 &62.3 &\textbf{\textcolor{blue}{68.2}} &59.7 &58.6 &54.3 &49.2 &32.5 \\
\hline
\multirow{3}{*}{speed ($fps$)} &GPU &195 &- &128 &83 &240 &50 &63   &41 &40 &47 &77 &56 \\
&CPU &45 &47 &41 &18 &15 &7 &5 &5 &5 &6 &10 &4 \\
&AGX &55 &- &36 &22 &39 &13 &13 &10 &10 &11 &17 &11 \\
\hline
\end{tabular}}
\end{center}
\end{table*}

{\noindent \textbf{TrackingNet. }}
TrackingNet~\cite{trackingnet} is a large-scale dataset containing diverse object categories and scenes. 
Its test set has 511 video sequences.
We upload our tracker's results to TrackingNet's official evaluation server.
The AUC, P$_{Norm}$, and P are shown in Table~\ref{tab-sota}.
Our tracker has the best real-time results, which are 76.6\%, 82.6\%, and 72.9\% for AUC, P$_{Norm}$, and P, respectively.
Compared with the state-of-the-art lightweight tracker LightTrack, our method performs 4.1\%, 4.8\%, and 3.4\% higher in terms of AUC, P$_{Norm}$, and P, with better speed.
Compared with the popular tracker PrDiMP, our method outperforms it by 0.8\%, 1\% and 2.5\% for AUC, P$_{Norm}$, and P, with 4$\times$ higher speed on GPU, 7$\times$ higher speed on CPU, and 5$\times$ higher speed on AGX.

{\noindent \textbf{LaSOT. }}
LaSOT~\cite{LaSOT} is a large-scale long-term dataset containing 1120 videos for training and 280 videos for testing.  
We follow the one-pass evaluation to test different tracking methods on the LaSOT test set.
The Success (AUC) and Precision (P$_{Norm}$ and P) scores are shown in Table~\ref{tab-sota}.
Our tracker has the best real-time results, which are 59.3\%, 68.7\%, and 61.0\% for AUC, P$_{Norm}$, and P, respectively.
Compared with the recent lightweight tracker LightTrack, our method outperforms it by 5.5\% and 7.3\% for AUC and P, with better speed.
Compared with the popular tracker PrDiMP, our tracker offers competitive performance (59.3 $vs.$ 59.8 in AUC) but with much faster speed.

{\noindent \textbf{GOT-10k. }}
GOT-10k~\cite{GOT10K} is a large-scale dataset containing a wide range of challenges. 
We submit HCAT's results to the official server.
The obtained AO and SR$_{T}$ scores are shown in Table~\ref{tab-sota}.
Compared with LightTrack, our method outperforms it by 4.0\% and 5.5\% for AO and SR$_{0.5}$, with better speed.
Compared with the popular tracker PrDiMP, our tracker outperforms it by 1.7\%, 2.7\%, and 2.4\% higher for AO, SR$_{0.5}$, and SR$_{0.75}$, with much faster speed.

\subsection{Evaluation on VOT2020 Datasets}
We compare our tracker with some state-of-the-art trackers on the challenging dataset VOT2020~\cite{vot2020} . 
VOT2020 contains 60 challenging videos with mask annotation. 
VOT2020 adopts EAO (expected average overlap) as the final metric to measure the robustness and accuracy of trackers. 
As our tracker does not generate masks, we only compare trackers that submit bounding boxes. 
The results are shown in Table~\ref{tab-sota-vot}.
Our tracker has the best real-time results.
Our tracker's performance is comparable to the powerful tracker STARK-S50, only 0.4\% lower in EAO, and 1.9\% higher in Robustness, with 4$\times$ higher speed on GPU, 5$\times$ higher speed on CPU, and 3$\times$ higher speed on AGX.

\begin{table*}[!t]
\setlength{\abovecaptionskip}{0cm} 
\setlength{\belowcaptionskip}{-0.2cm}
\caption{State-of-the-art comparison on VOT2020. The best real-time results
are shown in \textbf{\textcolor{red}{red}} fonts, and the best non-real-time results are shown in \textbf{\textcolor{blue}{blue}} fonts.}
\label{tab-sota-vot}
\begin{center}
\resizebox{\linewidth}{!}{
\setlength{\tabcolsep}{4mm}{
\begin{tabular}{c|cc cc|ccc}
\hline
 &\multicolumn{4}{c|}{Real-time} &\multicolumn{3}{c}{Non-real-time} \\
\hline
&Ours 
&E.T.Track\cite{ETTrack}
&LightTrack\cite{yan2021lighttrack}
&ATOM\cite{ATOM}
&STARK-ST50\cite{Stark}
&STARK-S50\cite{Stark}
&DiMP\cite{DiMP} \\
\hline
EAO &\textbf{\textcolor{red}{27.6}} &26.7 &24.2 &27.1 &\textbf{\textcolor{blue}{30.8}} &28.0 &27.4 \\
Accuracy &45.5 &43.2 &42.2 &\textbf{\textcolor{red}{46.2}} &\textbf{\textcolor{blue}{47.8}} &47.7 &45.7 \\
Robustness &\textbf{\textcolor{red}{74.7}} &74.1 &68.9 &73.4 &\textbf{\textcolor{blue}{79.9}} &72.8 &74.0 \\
\hline
GPU speed ($fps$) &195 &- &128 &83 &50 &50   &77   \\
CPU speed ($fps$) &45 &47 &41 &18 &7 &8   &10   \\
AGX speed ($fps$) &55 &- &36 &22 &13 &15   &17   \\
\hline
\end{tabular}}}
\end{center}
\end{table*}

\subsection{Evaluation on Other Datasets}
We evaluate our tracker on some common small-scale datasets, including
OTB100, NFS, and UAV123~\cite{OTB2015,NFS,UAV}. 
Small-scale datasets easily overfit, and many methods use different hyperparameters for each datasets. Our tracker uses the default hyperparameters without adjustment.
We report the AUC scores in Table~\ref{tab-sota-small}.
Our method outperforms LightTrack on these three datasets, with the best real-time results on NFS and the second best results on OTB100 and UAV123.

\begin{table*}[!t]
\setlength{\abovecaptionskip}{0cm} 
\setlength{\belowcaptionskip}{-0.2cm}
\caption{State-of-the-art comparison on OTB100, NFS, and UAV123 in terms of AUC score. The best real-time results 
are shown in \textbf{\textcolor{red}{red}} fonts, and the best non-real-time results are shown in \textbf{\textcolor{blue}{blue}} fonts.}
\label{tab-sota-small}
\begin{center}
\resizebox{\linewidth}{!}{
\begin{tabular}{c|cc ccc|c ccc cc}
\hline
 &\multicolumn{5}{c|}{Real-time} &\multicolumn{6}{c}{Non-real-time} \\
\hline
&Ours 
&E.T.Track~\cite{ETTrack}
&LightTrack\cite{yan2021lighttrack}
&ATOM\cite{ATOM} &ECO\cite{ECO}
&TransT\cite{TransT}
&TrDimp\cite{Wang_2021_Transformer} &TrSiam\cite{Wang_2021_Transformer}
&PrDiMP\cite{PrDiMP} &DiMP\cite{DiMP} &SiamRPN++\cite{SiamRPNplusplus}  \\
\hline
OTB100 &68.1 &67.8 &66.2 &66.9 &\textbf{\textcolor{red}{69.1}} &69.4 &\textbf{\textcolor{blue}{71.1}} &70.8 &69.6 &68.4 &69.6 \\
NFS &\textbf{\textcolor{red}{63.5}} &59.0 &55.3 &58.4 &46.6 &65.7 &\textbf{\textcolor{blue}{66.5}} &65.8 &63.5 &62.0 &50.2 \\
UAV123 &62.7 &62.3 &62.5 &\textbf{\textcolor{red}{64.2}} &53.2 &\textbf{\textcolor{blue}{69.1}} &67.5 &67.4 &68.0 &65.3 &61.6 \\
\hline
GPU speed ($fps$) &195 &- &128 &83 &240 &63   &41 &40 &47 &77 &56 \\
CPU speed ($fps$) &45 &47 &41 &18 &15 &5 &5 &5 &6 &10 &4 \\
AGX speed ($fps$) &55 &- &36 &22 &39 &13 &10 &10 &11 &17 &11 \\
\hline
\end{tabular}}
\end{center}
\end{table*}

\subsection{Ablation Study and Analysis}
\label{sec-ablation}

{\noindent \textbf{Component-Wise Analysis. }}
Table~\ref{tab-component} shows the component-wise study results.
The model speed after ONNX acceleration (indicated as ONNX speed) is also reported in this table.
\#4 is our default model. 
In \#2, the baseline is our method without the hierarchical connection structure and the FS module.
In \#3, after employing the hierarchical connection, the tracker can achieve 0.8\% AO improvement on GOT-10k, 2.4\% AUC improvement on TrackingNet, and 1.6\% AUC improvement on LaSOT.
This result verifies the effectiveness of our hierarchical cross-attention manner.
In addition, we also apply hierarchical connection structure to the original TransT. We only replace the juxtaposed cross-attention in TransT with our hierarchical cross-attention; the others remain unchanged.
The results of TransT with our hierarchical cross-attention are shown in Table~\ref{tab-sota-transt}, denoted as TransT\_H. Our method improves TransT to the level of state-of-the-art performance. TransT\_H has second best performance on GOT-10k, best performance on TrackingNet, and second best performance on LaSOT.
Except for LaSOT, all the results are better than STARK-ST50 with the same backbone network. STARK-ST50 is an online update method, while our TransT\_H is a completely offline method.

In \#4, after employing the FS module, the change in performance is negligible, but the speedup is noticeable on both CPU and AGX. 
The FS module reduces the computational amount by sparsing the template features, resulting in 12.5\% and 10\% speedup of the PyTorch model on CPU and AGX.
Given the redundant template feature, sparsification has no negative impact on performance.
We notice a slight slowdown on the GPU for the tracker using the FS module.
As we discussed in Section~\ref{sec-int} and Section~\ref{sec-method}, the computational amount of single operation layer is not the bottleneck of the speed on GPU, but the serial layer introduced by FS module slows down the model slightly.
However, as the speed is extremely high on the GPU, this speed reduction is negligible.
For the ONNX model, the FS module improves the speed on three platforms.

In \#1, the baseline-SA is the model that replaces the cross-attention layers in the baseline with self-attention layers.
The experimental results show that self-attention is not as effective as cross-attention for our model.
Therefore, we discard self-attention and adopt a full cross-attention design.

\begin{table*}[!t]
\setlength{\abovecaptionskip}{0cm} 
\setlength{\belowcaptionskip}{-0.2cm}
\caption{Component-wise study. The best results 
are shown in \textbf{\textcolor{red}{red}} fonts.}
\label{tab-component}
\begin{center}
\resizebox{\linewidth}{!}{
\setlength{\tabcolsep}{1mm}{
\begin{tabular}{c|c|ccc|ccc|ccc|ccc|ccc}
\hline
\multirow{2}{*}{\#} 
&\multirow{2}{*}{Component}
&\multicolumn{3}{c|}{GOT-10k}	
&\multicolumn{3}{c|}{TrackingNet}	&\multicolumn{3}{c|}{LaSOT}
&\multicolumn{3}{c|}{PyTorch Speed ($fps$)}
&\multicolumn{3}{c}{ONNX Speed ($fps$)}\\
\cline{3-17}
&  
&AO	    &SR$_{0.5}$	    &SR$_{0.75}$
&AUC    &P$_{Norm}$    &P	
&AUC	&P$_{Norm}$    &P	
&GPU	&CPU    &AGX
&GPU	&CPU    &AGX\\
\hline
1	&Baseline-SA   	
&60.1	&71.3	&49.1\
&74.0	&79.8	&68.8\
&54.2	&62.6	&54.0\
&192	&54	&63\
&572	&77	&119\\
2	&Baseline   	
&64.7	&76.6	&54.1\
&74.2	&80.6	&70.3\
&57.4	&67.2	&58.1\
&197	&40	&50\
&575	&80	&110\\

3	&+Hierarchical	
&\textbf{\textcolor{red}{65.5}}	&\textbf{\textcolor{red}{76.7}}	&\textbf{\textcolor{red}{57.0}}\
&\textbf{\textcolor{red}{76.6}}	&82.3	&72.6\
&59.0	&68.5	&60.6\
&197	&40	&50\
&575	&80	&110\\
4	&+FS Module
&65.1	&76.5	&56.7\
&\textbf{\textcolor{red}{76.6}}	&\textbf{\textcolor{red}{82.6}}	&\textbf{\textcolor{red}{72.9}}\
&\textbf{\textcolor{red}{59.3}}	&\textbf{\textcolor{red}{68.7}}	&\textbf{\textcolor{red}{61.0}}\
&195	&45	&55\
&589	&90	&127\\
\hline
\end{tabular}}}
\end{center}
\end{table*}

\begin{table*}[!t]
\setlength{\abovecaptionskip}{0cm} 
\setlength{\belowcaptionskip}{-0.2cm}
\caption{State-of-the-art comparison of TransT\_H. The best two results 
are shown in \textbf{\textcolor{red}{red}} fonts and \textbf{\textcolor{blue}{blue}} fonts.}
\label{tab-sota-transt}
\begin{center}
\resizebox{\linewidth}{!}{
\setlength{\tabcolsep}{4mm}{
\begin{tabular}{c|ccc|ccc|ccc}
\hline
\multirow{2}{*}{Method}
&\multicolumn{3}{c|}{GOT-10k}	
&\multicolumn{3}{c|}{TrackingNet}	&\multicolumn{3}{c}{LaSOT}\\
\cline{2-10}
&AO	    &SR$_{0.5}$	   &SR$_{0.75}$
&AUC    &P$_{Norm}$    &P	
&AUC	&P$_{Norm}$    &P	
\\
\hline
DiMP\cite{DiMP}
&61.1	&71.7	&49.2
&74.0	&80.1	&68.7	
&56.9	&65.0	&56.7
\\
SiamPRN++\cite{SiamRPNplusplus}
&51.7	&61.6	&32.5
&73.3	&80.0	&69.4
&49.6	&56.9	&49.1
\\
PrDiMP\cite{PrDiMP}
&63.4	&73.8	&54.3
&75.8	&81.6	&70.4
&59.8	&68.8	&60.8
\\
Ocean\cite{Ocean}	
&61.1	&72.1	&47.3
&-	&-	&-	
&56.0	&65.1	&56.6
\\
SiamR-CNN\cite{SiamRCNN}   
&64.9	&72.8	&59.7
&81.2	&85.4	&80.0
&64.8	&72.2	&-
\\
STMTrack\cite{fu2021stmtrack}  	
&64.2	&73.7	&57.5
&80.3	&85.1	&76.7
&60.6	&69.3	&63.3
\\
TrSiam\cite{Wang_2021_Transformer}  
&67.3	&78.7	&58.6
&78.1	&82.9	&72.7		
&62.4	&-	&60.0\\
TrDiMP\cite{Wang_2021_Transformer} 
&68.8	&80.5	&59.7
&78.4	&83.3	&73.1
&63.9	&-	&61.4\\
ARDiMPsuper\cite{Alpha-Refine}  	
&70.1	&80.0	&64.2 
&80.5	&85.6	&78.3
&65.3	&73.2	&68.0\\
DualTFR\cite{DualTFR}
&\textbf{\textcolor{red}{73.5}}	&\textbf{\textcolor{red}{84.8}}	&\textbf{\textcolor{red}{69.9}}
&80.1	&84.9	&-	
&63.5	&72.0	&66.5\\
DTT\cite{DTT}
&68.9	&79.8	&62.2
&79.6	&85.0	&78.9	
&60.1	&-	&-\\
STARK-S50\cite{Stark}  
&67.2	&76.1	&61.2
&80.3	&85.1	&-		
&65.8	&-	&-\\
STARK-ST50\cite{Stark}  
&68.0	&77.7	&62.3
&81.3	&86.1	&-		
&\textbf{\textcolor{red}{66.6}}	&-	&-\\
TransT\cite{TransT}	
&72.3	&\textbf{\textcolor{blue}{82.4}}	&68.2\
&\textbf{\textcolor{blue}{81.4}}	&\textbf{\textcolor{blue}{86.7}}	&\textbf{\textcolor{blue}{80.3}}\
&64.9	&\textbf{\textcolor{blue}{73.8}}	&\textbf{\textcolor{blue}{69.0}}\\
TransT\_H
&\textbf{\textcolor{blue}{72.4}}	&82.0	&\textbf{\textcolor{blue}{68.5}}\
&\textbf{\textcolor{red}{82.2}}	&\textbf{\textcolor{red}{87.0}}	&\textbf{\textcolor{red}{80.4}}\
&\textbf{\textcolor{blue}{66.2}}	&\textbf{\textcolor{red}{75.1}}	&\textbf{\textcolor{red}{70.7}}\\
\hline
\end{tabular}}}
\end{center}
\end{table*}

{\noindent \textbf{Number of Sparse Vectors. }}
We explore the effect of the number of sparse vectors. 
In Table~\ref{tab-s}, $S$ indicates the number of sparse vectors.
As $S$ increases, the performance gradually increases.
When $S=16$, the performance is at par with the performance without the FS module.
This result verifies the occurrence of redundancy in the template feature vectors.
A total of 16 sparse vectors can achieve performance similar to the original 64 template vectors.

{\noindent \textbf{Number of HCA Layers. }}
We explore the effect of the number of HCA layers. 
In the results shown in Table~\ref{tab-N}, $N$ indicates the number of HCA layers.
When $N=1$, the model is extremely fast, the ONNX model can reach 702 $fps$ on GPU, 104 $fps$ on CPU, and 147 $fps$ on AGX, with good performance.
We believe that the extremely fast tracker facilitates many vision tasks and real-world applications.

\begin{table*}[!t]
\begin{minipage}[c]{0.34\textwidth}
\caption{Ablation study with different numbers of template queries $S$. The best results 
are shown in \textbf{\textcolor{red}{red}} fonts.}
\label{tab-s}
\end{minipage}
\begin{minipage}[c]{0.64\textwidth}
\setlength{\tabcolsep}{2mm}{
\resizebox{\textwidth}{!}{
\begin{tabular}{c|ccc|ccc|ccc}
\hline
\multirow{2}{*}{$S$}
&\multicolumn{3}{c|}{GOT-10k}	
&\multicolumn{3}{c|}{TrackingNet}	&\multicolumn{3}{c}{LaSOT}\\
\cline{2-10}
&AO	    &SR$_{0.5}$	   &SR$_{0.75}$
&AUC    &P$_{Norm}$    &P	
&AUC	&P$_{Norm}$    &P	
\\
\hline
1   	
&62.3	&73.7	&51.4\
&74.7	&80.5	&70.1\
&57.6	&67.0	&58.3\\
4	
&64.5	&75.3	&55.7\
&75.8	&81.4	&71.4\
&58.3	&67.4	&58.8\\
9	
&63.9	&75.2	&55.6\
&76.2	&81.7	&71.9\
&58.3	&67.5	&59.1\\
16	
&\textbf{\textcolor{red}{65.1}}	&\textbf{\textcolor{red}{76.5}}	&\textbf{\textcolor{red}{56.7}}\
&\textbf{\textcolor{red}{76.6}}	&\textbf{\textcolor{red}{82.6}}	&\textbf{\textcolor{red}{72.9}}\
&59.3	&68.7	&61.0\\
25	
&64.5	&75.9	&56.1\
&76.2	&82.0	&72.2\
&\textbf{\textcolor{red}{60.7}}	&\textbf{\textcolor{red}{70.1}}	&\textbf{\textcolor{red}{61.9}}\\
\hline
\end{tabular}}}
\end{minipage}
\end{table*}

\begin{table*}[!t]
\setlength{\abovecaptionskip}{0cm} 
\setlength{\belowcaptionskip}{-0.2cm}
\caption{Ablation study with different numbers of feature fusion layers $N$. The best results 
are shown in \textbf{\textcolor{red}{red}} fonts.}
\label{tab-N}
\begin{center}
\resizebox{\linewidth}{!}{
\setlength{\tabcolsep}{1.6mm}{
\begin{tabular}{c|ccc|ccc|ccc|ccc|ccc}
\hline
\multirow{2}{*}{$N$}
&\multicolumn{3}{c|}{GOT-10k}	
&\multicolumn{3}{c|}{TrackingNet}	
&\multicolumn{3}{c|}{LaSOT}
&\multicolumn{3}{c|}{PyTorch Speed ($fps$)}
&\multicolumn{3}{c}{ONNX Speed ($fps$)}\\
\cline{2-16}
&AO	    &SR$_{0.5}$	    &SR$_{0.75}$
&AUC    &P$_{Norm}$    &P	
&AUC	&P$_{Norm}$    &P	
&GPU	&CPU    &AGX
&GPU	&CPU    &AGX\\
\hline
1 	
&62.4	&73.8	&51.9\
&74.2	&80.0	&69.3\
&57.6	&67.2	&57.7\
&240	&46	&69\
&702	&104	&147\\
2 	
&\textbf{\textcolor{red}{65.1}}	&\textbf{\textcolor{red}{76.5}}	&\textbf{\textcolor{red}{56.7}}\
&\textbf{\textcolor{red}{76.6}}	&\textbf{\textcolor{red}{82.6}}	&\textbf{\textcolor{red}{72.9}}\
&\textbf{\textcolor{red}{59.3}}	&\textbf{\textcolor{red}{68.7}}	&\textbf{\textcolor{red}{61.0}}\
&195	&45	&55\
&589	&90	&127\\
\hline
\end{tabular}}}
\end{center}
\end{table*}

\section{Conclusion}
In this work, we propose an efficient and accurate tracking framework based on a novel feature fusion network. 
The feature fusion network is composed of a feature sparsification module and a hierarchical cross-attention transformer.
The feature sparsification module sparses the template features to reduce the computational amount of the transformer.
The hierarchical cross-attention transformer employs a full cross-attention design and a shallow structure to improve efficiency, and it also employs the hierarchical connection structure to enhance the representation ability.
The experimental results on many benchmarks indicate that our tracker outperforms the state-of-the-art high-speed methods.
The PyTorch model runs at 195 $fps$ on GPU, 45 $fps$ on CPU, and 55 $fps$ on the edge AI platform NVidia Jetson AGX Xavier, and the ONNX model runs at 589 $fps$ on GPU, 90 $fps$ on CPU, and 127 $fps$ on NVidia Jetson AGX Xavier. 

{\noindent \textbf{Acknowledgement. }}This work was supported in part by the National Natural Science Foundation of China (NSFC) under Grant 61902420 and 62022021, in part by Joint Fund of Ministry of Education for Equipment Pre-research under Grant 8091B032155, in part by National Defense Basic Scientific Research Program under Grant WDZC20215250205, in part by the Science and Technology Innovation Foundation of Dalian under Grant no. 2020JJ26GX036, and in part by the Fundamental Research Funds for the Central Universities under Grant DUT21LAB127.

\clearpage

\bibliographystyle{splncs04}
\bibliography{egbib}
\end{document}